\documentclass[10pt,twocolumn,letterpaper]{article}

\usepackage{iccv}
\usepackage{times}
\usepackage{epsfig}
\usepackage{graphicx}
\usepackage{amsmath}
\usepackage{amssymb}
\usepackage{graphicx}
\usepackage{amsmath}
\usepackage{amssymb}
\usepackage{booktabs}
\usepackage{cite}
\usepackage{booktabs}
\usepackage{multirow}
\usepackage[accsupp]{axessibility}
\usepackage[table,xcdraw]{xcolor}

\usepackage[pagebackref=true,breaklinks=true,letterpaper=true,colorlinks,bookmarks=false]{hyperref}

\iccvfinalcopy 

\def\iccvPaperID{3658} 

\ificcvfinal\fi

\begin{document}

\title{Scene Matters: Model-based Deep Video Compression}

\author{Lv Tang \quad Xinfeng Zhang\thanks{Corresponding author. This work was supported by the National Natural Science Foundation under Grant 62071449 and U20A20184, and the Fundamental Research Funds for the Central Universities.} \quad Gai Zhang\quad Xiaoqi Ma \\
University of Chinese Academy of Sciences, Beijing, China  \\
{\tt\small luckybird1994@gmail.com}, {\tt\small xfzhang@ucas.ac.cn}, {\tt\small zhanggai16@mails.ucas.ac.cn},
{\tt\small maxiaoqi197@gmail.com}
}

\maketitle
\ificcvfinal\thispagestyle{empty}\fi

\begin{abstract}
Video compression has always been a popular research area, where many traditional and deep video compression methods have been proposed. These methods typically rely on signal prediction theory to enhance compression performance by designing high efficient intra and inter prediction strategies and compressing video frames one by one. In this paper, we propose a novel model-based video compression (MVC) framework that regards scenes as the fundamental units for video sequences. Our proposed MVC directly models the intensity variation of the entire video sequence in one scene, seeking non-redundant representations instead of reducing redundancy through spatio-temporal predictions. To achieve this, we employ implicit neural representation as our basic modeling architecture. To improve the efficiency of video modeling, we first propose context-related spatial positional embedding and frequency domain supervision  in spatial context enhancement. For temporal correlation capturing, we design the scene flow constrain mechanism and temporal contrastive loss. Extensive experimental results demonstrate that our method achieves up to a 20\% bitrate reduction compared to the latest video coding standard H.266 and is more efficient in decoding than existing video coding strategies.
\end{abstract}

\section{Introduction} \label{sec:intro}
Recently, videos have become ubiquitous in people's daily lives, from short-form videos to conference and surveillance videos. Efficiently storing and transmitting video data has become a significant challenge due to the vast amounts and explosive growth of such data. To address this challenge, multiple video compression standards have been developed based on traditional hybrid video coding frameworks, such as H.264/AVC~\cite{DBLP:journals/tcsv/WiegandSBL03}, H.265/HEVC~\cite{DBLP:journals/tcsv/SullivanOHW12}, and H.266/VVC~\cite{bross2021overview}, as well as deep-learning based video compression (DLVC) methods~\cite{DBLP:conf/dcc/CuiZZJZWZ17,DBLP:journals/tcsv/YanLLLLW19,DBLP:conf/cvpr/LuO0ZCG19,DBLP:conf/cvpr/Lin0L020,DBLP:conf/cvpr/HuL021,li2021deep,ho2022canf}.

\begin{figure}
\centering
\includegraphics[width=\linewidth]{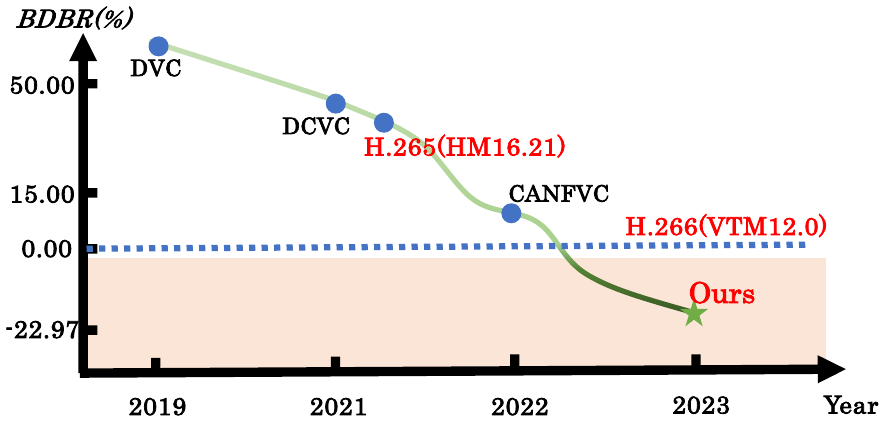}
\caption{BDBR(\%)~\cite{bjontegaard2001calculation} performances of different methods when compared with H.266 on the real-world surveillance video sequences in terms of PSNR. DVC~\cite{DBLP:conf/cvpr/LuO0ZCG19}, DCVC~\cite{li2021deep} and CANFVC~\cite{ho2022canf} are three DLVC methods.}
\label{introd}
\vspace{-0.5cm}
\end{figure}

Both traditional hybrid video coding frameworks and existing deep learning-based video compression (DLVC) methods follow the same approach of compressing videos by designing various technique modules to reduce spatial and temporal redundancy. In traditional hybrid video coding, each video frame is divided into blocks, and intra- and inter-prediction techniques~\cite{zhang2018improved,DBLP:journals/tip/WangWZWM19} are used to reduce spatial and temporal redundancy. On the other hand, DLVC methods~\cite{DBLP:conf/cvpr/LuO0ZCG19,DBLP:conf/cvpr/Lin0L020,DBLP:conf/cvpr/HuL021,ho2022canf}, unlike traditional compression methods, use neural networks to design end-to-end intra- and inter-prediction modules for the entire frame. Despite the careful design of these techniques, both traditional and DLVC methods compress a video sequence progressively in a block-by-block or frame-by-frame style and only use neighboring pixels in the same frame or neighboring frames as reference to derive intra- or inter-prediction values. Since video sequences are captured at high framerates, such as 30fps or 60fps, the same scene may appear in hundreds of frames that are highly correlated in the temporal domain. However, existing compression strategies are not well-equipped to remove scene redundancy in the block- or frame-level prediction. {As demonstrated in Fig. \ref{introd}, the performance of existing state-of-the-art (SOTA) DLVC methods still lags behind that of the traditional H.266 standard.}

\begin{figure}
\centering
\includegraphics[scale=0.81]{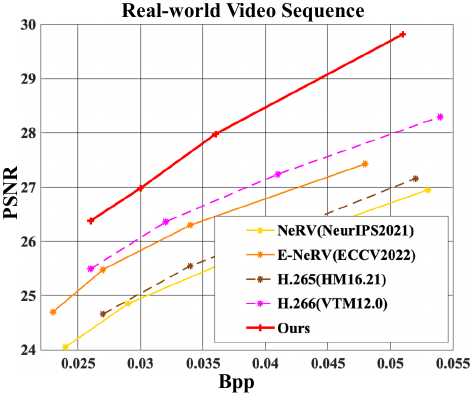}
\caption{The performance of existing SOTA video INR methods when applied to the video compression task.}
\label{introd2}
\vspace{-0.5cm}
\end{figure}

To overcome the performance bottleneck in video compression, this paper proposes an innovative video coding paradigm that seeks to find a compact subspace for a video sequence of the same scene, rather than reducing spatio-temporal redundancy through block-level or frame-level prediction methods. This paradigm replaces explicit redundancy reduction through local prediction with implicit compact subspace modeling for the entire scene. {  Consequently, finding a suitable modeling tool to represent the scene is crucial to this paradigm.} Recently, implicit neural representation (INR) has gained popularity for its strong ability to model a wide variety of signals by a deep network. INR has been already applied to various tasks to represent different objects, such as RGB images~\cite{sitzmann2020implicit}, 3D shapes~\cite{park2019deepsdf,sitzmann2020implicit} and scenes~\cite{DBLP:conf/eccv/MildenhallSTBRN20,li2022neural}. Considering that original signals can be implicitly encoded in network’s parameters, some researchers apply the INR to the image compression task~\cite{DBLP:journals/corr/abs-2112-04267,dupont2021coin,DBLP:journals/corr/abs-2201-12904}, and achieve competitive performance compared to traditional image compression method JPEG2000. Since the compressed bitstream is network’s parameters, these image compression methods can be regarded as model-based image compression. Due to these characteristics, INR is a promising candidate for the backbone network of the proposed video compression paradigm.

In contrast to model-based image compression, model-based video compression (MVC) is barely explored. In MVC, sequence modeling is an extra significant factor, which is the main challenge for video compression. However, the representation ability of the primal video INR methods~\cite{DBLP:journals/corr/abs-2207-08132,chen2021nerv} is limited. If we directly apply these methods to the video compression task, NeRV, is even inferior to traditional video coding standard H.265~\cite{DBLP:journals/tcsv/SullivanOHW12}, as shown in Fig. \ref{introd2}. This demonstrates that existing SOTA INR methods are unable to achieve higher-quality reconstruction results when given limited network parameters, highlighting the potential for further developments in applying video INR to video compression tasks. In this paper, we further improve the sequence modeling ability of video INR in spatial context enhancement and temporal correlation capturing.

In spatial context capturing, existing video INR methods, such as those presented in \cite{DBLP:journals/corr/abs-2207-08132,chen2021nerv,DBLP:journals/corr/abs-2208-03742}, use a learnable network to reconstruct video frames from context-agnostic spatial positional embeddings. However, to handle spatial variations between different frames and achieve higher-quality reconstruction results, these methods typically require additional network parameters (\textit{bitrates}), which can adversely impact rate-distortion (RD) performance. To address this issue, we propose a context-related spatial positional embedding (CRSPE) method in this paper. Additionally, some works \cite{DBLP:conf/cvpr/LeeJ22,DBLP:journals/corr/abs-2207-01831} have proposed frequency-aware operations in their networks to improve the context capturing ability and capture high-frequency image details. However, these operations often come with an added cost of network parameters that can degrade compression performance. To address this problem and maintain a balance between compression performance and reconstruction quality, we introduce a frequency domain supervision (FDS) module that can capture high-frequency details without requiring additional bitrates.

Temporal correlation is a critical factor for INR methods to improve the representation efficiency of different frames. Existing video INR methods primarily rely on different time positional encodings to distinguish between frames and expect the network to implicitly learn temporal correlation. While these encodings can capture temporal correlation to some extent, they struggle to explore complex temporal correlations, particularly for long video sequences. To address this limitation, we introduce a scene flow constraint mechanism (SFCM) for short-term temporal correlation and a temporal contrastive loss (TCL) for long-term temporal correlation in this paper. These mechanisms do not increase network parameters and are well-suited for the MVC task. As illustrated in Fig. \ref{introd}, our proposed framework already outperforms H.266 \cite{bross2021overview} significantly, indicating the potential of MVC methods. Our main contributions are:

\begin{figure*}[!hbpt]
\centering
\includegraphics[width=\linewidth]{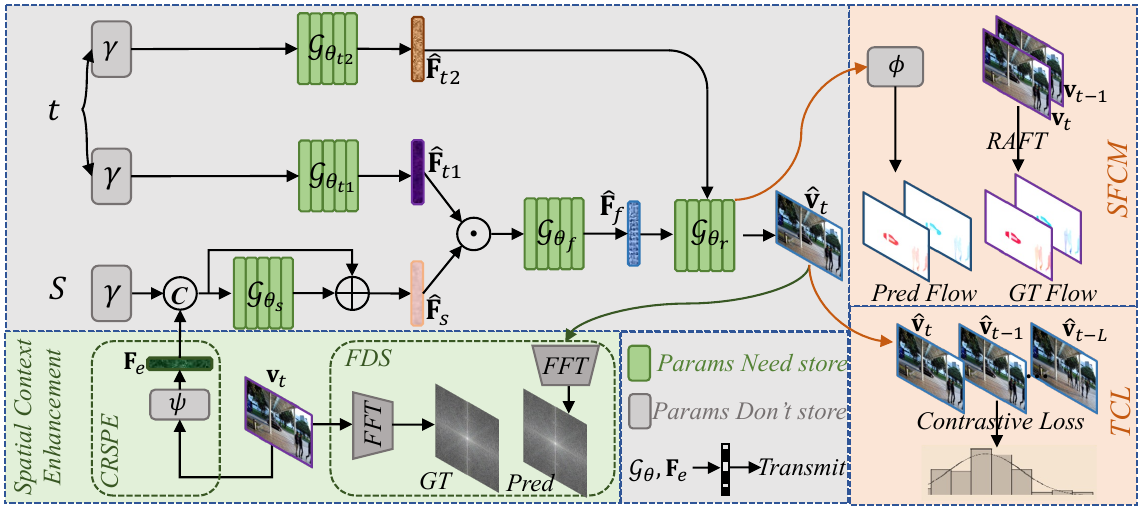}
\caption{The framework of our proposed model-based video compression method.}
\label{framework}
\end{figure*}

\begin{itemize}

\item We propose an MVC that seeks to identify more compact sub-spaces for video sequences. Unlike existing methods that rely on explicit spatio-temporal redundancy reduction through signal prediction at the block or frame level, our framework uses the correlations between all frames in a video scene simultaneously. 

\item { To address the limitations of existing video INR methods when applied to video compression}, we introduce CRSPE and FDS in spatial context enhancement, which can handle spatial variations of different frames and capture high-frequency details. We further design SFCM and TCL for temporal correlation modeling.

\item Extensive experiments are conducted on different databases, and detailed analyses are provided for our designed modules. Experimental results show that our proposed method can outperform H.266 (VTM12.0), which demonstrates the superiority of our proposed method and may inspire researchers to explore video compression in a new light.

\end{itemize}

\section{Related works}
\subsection{Video Compression Methods.} 
Traditional video coding methods mainly follow hybrid video coding framework by improving various coding tools~e.g., short-distance intra prediction~\cite{cao2012short}, affine motion estimation~\cite{zhang2018improved} and low-rank in-loop filter\cite{zhang2016low}. However, these coding tools are designed so delicately that the traditional coding framework faces performance bottleneck~\cite{DBLP:journals/tcsv/ZhangZFMCK20}. Deep learning technique has also been applied to improve video coding performance, and achieved great progress in recent years. According to recent review works in ~\cite{DBLP:journals/csur/LiuLLLW20,DBLP:journals/tcsv/MaZJZWW20}, deep learning based video compression methods can be roughly divided into two categories: deep-tool methods and deep-framework methods. Deep-tool methods~\cite{DBLP:conf/vcip/ChenLS0CM17,DBLP:journals/tip/LiuYGCJW16,DBLP:conf/vcip/SongLLW17,DBLP:conf/eccv/LuOXZGS18,DBLP:conf/cvpr/YangXWL18,zhao2019enhanced} still follow traditional hybrid video coding framework, and replace one or more coding tools by deep neural networks (DNN), such as intra/inter prediction, probability distribution prediction and in-loop filtering, in the traditional framework. However, restricted by the traditional framework, these methods cannot make full use of DNN, which limits its potential improvements. Hence, recent deep-framework methods~\cite{DBLP:conf/eccv/WuSK18,DBLP:conf/iccv/DjelouahCSS19,DBLP:conf/sips/PessoaATF20,DBLP:conf/iccv/HabibianRTC19,DBLP:conf/icassp/TangLWXD21,DBLP:conf/bmvc/Zhong0TTD21,DBLP:conf/iccv/RippelATNLB21,DBLP:conf/cvpr/YangMGT20,DBLP:journals/tip/YilmazT22,DBLP:journals/corr/abs-2104-14729,DBLP:conf/cvpr/LuO0ZCG19,DBLP:conf/accv/Tang020,DBLP:conf/ijcai/0061STSS19,DBLP:conf/icassp/LiSTH19,DBLP:conf/cvpr/Lin0L020,DBLP:journals/tip/0115TKSD22,DBLP:conf/cvpr/HuL021,DBLP:journals/tcsv/TangLKSD22,DBLP:conf/iccv/RippelNLBAB19,DBLP:conf/cvpr/AgustssonMJBHT20,DBLP:journals/pami/LuZO0G021,DBLP:conf/eccv/HuCXLOG20,DBLP:conf/iccv/TangLZDS21,DBLP:conf/iclr/YangYMM21,DBLP:journals/jstsp/YangMGT21,DBLP:conf/cvpr/ZhongLTKWD22,Hu_2022_CVPR,DBLP:journals/corr/abs-2207-14678,DBLP:conf/wacv/GhousePSWP23,DBLP:conf/eccv/ShiGWM22,DBLP:journals/corr/abs-2206-07307} propose to construct end-to-end DLVC frameworks. Although DLVC methods can leverage the benefits of end-to-end learning strategy, these methods still remove temporal redundancy by only referring to one or limited neighboring frames, which limits its performance improvement. In this paper, our proposed MVC method can simultaneously leverage the correlations of all frames in the scene and find a much more compact sub-space for videos.

\subsection{Implicit Neural Representation.}
INR is a new paradigm to parameterize a wide range of signals, and its key idea is to represent an object as a function approximated by neural networks. DeepSDF~\cite{DBLP:conf/cvpr/ParkFSNL19} is one of the early works on INR, which is a neural network representation for 3D shapes. Recently, many works are proposed to represent different objects with INR, such as 3D shapes~\cite{DBLP:conf/cvpr/MeschederONNG19} and scene representation~\cite{DBLP:conf/eccv/MildenhallSTBRN20,li2022neural}, etc. Due to the representation ability of INR, some works~\cite{DBLP:journals/corr/abs-2112-04267,dupont2021coin,DBLP:journals/corr/abs-2201-12904} introduce INR to image compression task, proposing MIC methods. The work INVC~\cite{DBLP:journals/corr/abs-2112-11312} also applies INR to video compression network, and it directly uses INR to represent each frame then estimates motion and residual information. In fact, the INVC still follows the framework of existing common deep video compression and cannot fully explore the potential of INR. Therefore, its performance is even lower than H.264. In this paper, we propose the MVC method by designing novel modules for spatial context enhancement and temporal correlation capturing to simultaneously model video sequences in one scene, and the performance of our MVC method can outperform H.266.

\section{Method}
We first introduce the preliminary knowledge of this paper. Then we elaborate on our main contributions in this paper. Finally, we describe the whole compression pipeline of this work. In this paper, the original video sequence containing $T$ frames is written as $\mathbf{V}=\{\mathbf{v}_{t}\}_{t=0}^{T-1}$, and the corresponding reconstruction video sequence is $\mathbf{\hat{V}}=\{\mathbf{\hat{v}}_{t}\}_{t=0}^{T-1}$. The framework of our proposed MVC network is in Fig. \ref{framework}.

\subsection{Preliminaries}
The typical video INR (V-INR) methods~\cite{DBLP:journals/corr/abs-2207-08132,chen2021nerv} represent the video as a mapping network $\mathcal{G}_\theta: \mathbb{R} \rightarrow \mathbb{R}^{3 \times H \times W}$ parameterized by the network weight $\theta$, where $H,W$ are the spatial size of the video frame. As shown in Fig. \ref{framework}, the whole mapping network contains five sub-networks: $\mathcal{G}_\theta=\{ \mathcal{G}_{\theta_{t1}},\mathcal{G}_{\theta_{t2}},\mathcal{G}_{\theta_{s}},\mathcal{G}_{\theta_{f}},\mathcal{G}_{\theta_{r}}\}$, and its corresponding weights are: $\theta=\{\theta_{t1},\theta_{t2},\theta_{s},\theta_{f},\theta_{r}\}$.

Firstly, the V-INR performs the regular positional encoding~\cite{DBLP:conf/eccv/MildenhallSTBRN20} on the scalar frame index $t$, then maps the temporal positional encoding to a feature vector $\hat{\mathbf{F}}_{t1} \in \mathbb{R}^d$ through the MLP operation $\mathcal{G}_{\theta_{t1}}$:
\begin{equation}
\begin{aligned}
\hat{\mathbf{F}}_{t1} = \mathcal{G}_{\theta_{t1}}(\gamma(t)),
\end{aligned}
\end{equation}
where the $\gamma(t)$ means the positional encoding:
\begin{equation}
\begin{aligned}
\gamma(t)=( &\sin \left(b^0 \pi t\right), \cos \left(b^0 \pi t\right), \ldots, \sin \left(b^{l-1} \pi t\right), \\ &\cos \left(b^{l-1} \pi t\right) ).
\end{aligned}
\label{positionalencoding}
\end{equation}
$b=1.25$ and $l=80$ follow the common setting in~\cite{DBLP:journals/corr/abs-2207-08132,chen2021nerv}.

Secondly, the V-INR initializes the normalized grid coordinates $S$, which are expected to contain the spatial context. The size of $S$ is $\mathbb{R}^{2 \times h \times w}$. For $S$, the V-INR first encodes it into $\hat{S}$ with similar positional encoding $\gamma(\cdot)$ in Eqn. \ref{positionalencoding}. Then, V-INR adopts a small transformer~\cite{DBLP:conf/nips/VaswaniSPUJGKP17} with single-head self-attention and residual connection to encourage the feature fusion among spatial locations:
\begin{equation}
\begin{aligned}
\hat{\mathbf{F}}_{s} = \mathcal{G}_{\theta_s}(\hat{S}) + \hat{S}, 
\end{aligned}
\end{equation}
where $\mathcal{G}_{\theta_s}$ is the transformer network. The ``\textit{C}" (Concatenation, Fig. \ref{framework}) dose not exist in the original V-INR network, which is the contribution proposed in this paper. 

Thirdly, the V-INR fuses $\hat{\mathbf{F}}_{t1}$ and $\hat{\mathbf{F}}_{s}$ with another MLP network $\mathcal{G}_{\theta_f}$:
\begin{equation}
\begin{aligned}
\hat{\mathbf{F}}_{f} = \mathcal{G}_{\theta_f}(\hat{\mathbf{F}}_{s} \odot \hat{\mathbf{F}}_{t1}). 
\end{aligned}
\end{equation}

Finally, the network $\mathcal{G}_{\theta_r}$ is used to reconstruct frames:
\begin{equation}
\begin{aligned}
\mathbf{\hat{v}}_{t} &= \mathcal{G}_{\theta_r}(\hat{\mathbf{F}}_{f},\hat{\mathbf{F}}_{t2}), where \\  
\hat{\mathbf{F}}_{t2} &= \mathcal{G}_{\theta_{t2}}(\gamma(t)).
\end{aligned}
\label{generation}
\end{equation}
In the reconstruction stage, the V-INR further leverages the MLP $\mathcal{G}_{\theta_{t2}}$ on $\gamma(t)$ to make sufficient and thorough use of the temporal embedding. Note that, $\mathcal{G}_{\theta_r}$ contains five up-sampling stages with pixel-shuffle operation. The  $\mathcal{L}_{1}$ loss is used for optimization:
\begin{equation}
\begin{aligned}
\mathcal{L}_{spa} = \mathcal{L}_{1}(\mathbf{\hat{v}}_{t}, \mathbf{v}_{t}).
\end{aligned}
\end{equation}

\begin{figure}[!t]
\centering
\includegraphics[width=\linewidth]{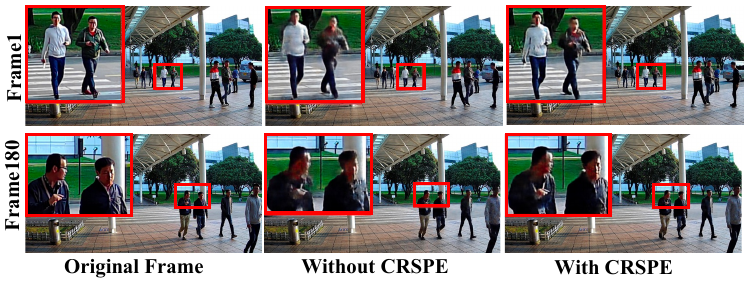}
\caption{Reconstructed results without/with CRSPE.}
\vspace{-0.5cm}
\label{Spatial}
\end{figure}

\subsection{Spatial Context Enhancement}

\subsubsection{Context-related Spatial Positional Embedding} 
Existing video INR methods implicitly represent spatial context by fixed grid coordinates $S$, which are context-agnostic spatial positional embeddings. However, they suffer from spatial variations among different frame contents, and the network would spend more encoding times and larger parameters. To address this problem, we propose the CRSPE. In Fig. \ref{Spatial}, we show two frames containing spatial variations in video sequence. With our proposed CRSPE, the network can reconstruct higher-quality results. 

For the original frame $\mathbf{v}_{t}$, we use a $80 \times 80$ convolutional operation $\psi$ to transform it to $\mathbf{F}_{e} \in \mathbb{R}^{c \times h \times w}$:
\begin{equation}
\begin{aligned}
\mathbf{F}_{e} = \psi(\mathbf{v}_{t}).
\end{aligned}
\end{equation}
In this paper, we set $c=3$ and the embedding $\mathbf{F}_{e}$ is the same spatial size as $S$.  Although transmitting the embedding $\mathbf{F}_e$ needs extra bitrates, the model can obtain better reconstruction quality, which would benefit the RD performance. Specifically, our proposed framework spends extra 10\% (Bits per pixel, Bpp) with about 0.9dB (PSNR) increasing for 720p videos.

\subsubsection{Frequency Domain Supervision} 
To further improve the performance of INR network, some works~\cite{DBLP:conf/cvpr/LeeJ22,DBLP:journals/corr/abs-2207-01831} utilized frequency-aware operations in their networks, which can capture high-frequency details of images. However, these operations are difficult to directly be applied to video compression task, since these operations need extra sophisticated modules and would introduce more coding bitrates. In order to keep more high-frequency detailed information, we propose the frequency-aware perceptual loss without adding network parameters. In particular, we utilize the fast Fourier transform (FFT) to transform $\mathbf{\hat{v}}_{t}$ and $\mathbf{v}_{t}$ into frequency domain, then calculate the $\mathcal{L}_{1}$ loss:
\begin{equation}
\begin{aligned}
\mathcal{L}_{freq} = \mathcal{L}_{1}(\mathbf{FFT}(\mathbf{\hat{v}}_{t}), \mathbf{FFT}(\mathbf{v}_{t})).
\end{aligned}
\end{equation}

\begin{figure}[!t]
\centering
\includegraphics[width=\linewidth]{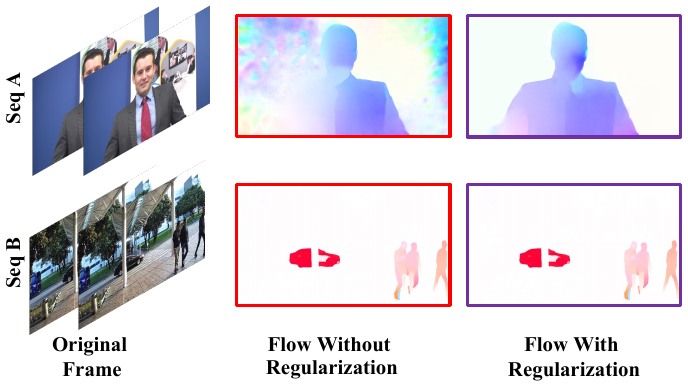}
\caption{The scene flow without/with regularization loss.}
\label{flow}
\vspace{-0.5cm}
\end{figure}

\subsection{Temporal Correlation Capturing}
Temporal correlation is another crucial factor for the MVC to distinguish the representation of different frames. Existing V-INR methods are not efficient in modeling complex temporal variations, especially with long-term temporal correlation, by only utilizing different positional encodings. Therefore, we first propose SFCM to capture short-term temporal correlation. We also design TCL to enhance the modeling efficiency for long-term temporal correlation.

\subsubsection{Scene Flow Constrain Mechanism.}

In Eqn. \ref{generation}, $\mathcal{G}_{\theta_r}(\cdot)$ has five up-sampling stages. The last stage contains the feature $\hat{\mathbf{Z}}_{t}$, which is for frame $\mathbf{\hat{v}}_{t}$ generation. In Fig. \ref{framework}, we design the extra scene flow prediction head $\phi$ on $\hat{\mathbf{Z}}_{t}$ to predict the forward and backward flows at timestamp $t$. $\mathbf{O}_{t}^{f}$ means the forward flow map from $t-1$ to $t$, and $\mathbf{O}_{t}^{b}$ is the backward flow map from $t$ to $t-1$. We do not evaluate the optical flow for $t=0$. To supervise the predicted flow maps, we estimate their corresponding ground truth (GT) $\{\mathbf{O}_{t}^{fgt},\mathbf{O}_{t}^{bgt}\}$ from original images $\{\mathbf{v}_{\mathbf{t-1}},\mathbf{v}_{t}\}$, through optical flow estimation algorithm RAFT~\cite{DBLP:conf/eccv/TeedD20}. Finally, the SFCM is optimized by $\mathcal{L}_{1}$ loss:
\begin{equation}
\begin{aligned}
\mathcal{L}_{vanilla-flow} = \mathcal{L}_{1}(\mathbf{O}_{t}^{f},\mathbf{O}_{t}^{fgt}) + \mathcal{L}_{1}(\mathbf{O}_{t}^{b},\mathbf{O}_{t}^{bgt}).
\end{aligned}
\end{equation}

Since we do not directly supervise the SFCM from the real annotation, the generated GT from RAFT would contain some noise due to different reasons, such as algorithm accuracy. As shown in Fig. \ref{flow}, the noisy GT-flow in the SeqA would disrupt the whole training process. To address this problem, we further design the regularization mechanism. Specifically, we use a $1\times1$ convolutional operation on $\hat{\mathbf{Z}}_{t}$ to evaluate the 2-channels regularization map $\mathbf{W}$, following by \textit{Softmax} operation. Finally, we select the values in the second channel, $\mathbf{W}^{(1)}$, to re-write the $\mathcal{L}_{flow}$:
\begin{equation}
\begin{aligned}
\mathcal{L}_{flow} = &\mathcal{L}_{1}(\mathbf{W}^{(1)} \cdot \mathbf{O}_{t}^{f}, \mathbf{W}^{(1)} \cdot \mathbf{O}_{t}^{fgt}) + \\ &\mathcal{L}_{1}(\mathbf{W}^{(1)} \cdot \mathbf{O}_{t}^{b}, \mathbf{W}^{(1)} \cdot \mathbf{O}_{t}^{bgt}).
\end{aligned}
\label{eqflow}
\end{equation}

Moreover, we further design the loss $\mathcal{L}_{ent}$ to let the $\mathbf{W}$ tend to be binary.
The $\mathcal{L}_{ent}$ is defined as:
\begin{equation}
\mathcal{L}_{ent} = - ( \mathbf{W}^{(0)} \cdot log (\mathbf{W}^{(0)}) + \mathbf{W}^{(1)} \cdot log (\mathbf{W}^{(1)}) ).
\label{ent}
\end{equation}
$\mathcal{L}_{ent}$ would be zero when the channels of $\mathbf{W}$ are the one-hot vector and would be maximum when they have an equal probability. As shown in Fig. \ref{flow}, the proposed regularization mechanism can erase some noisy flow information and not affect other flow information. 

\begin{figure}[!t]
\centering
\includegraphics[width=\linewidth]{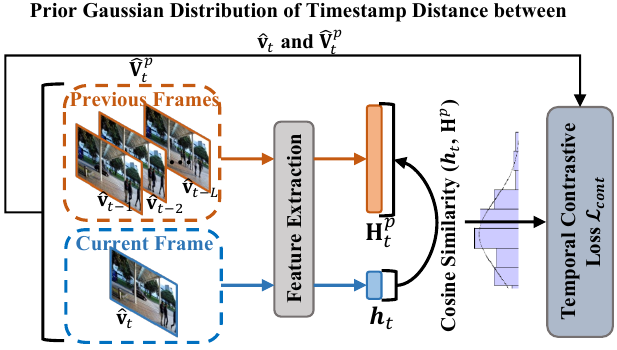}
\caption{Illustration of the proposed TCL. The prior Gaussian distribution is computed by the timestamp distance between current frame $\hat{\mathbf{v}}_t$ and the previous frames $\hat{\mathbf{V}}^p$.}
\label{Contrastive}
\vspace{-0.5cm}
\end{figure}

\subsubsection{Temporal Contrastive Loss}
Although SFCM can capture the short-term temporal correlation of two adjacent frames, the long-term temporal correlation is also important and has not been well utilized in previous video compression methods. To further improve the representation capability of our proposed network, we aim to model the long-term temporal correlation between the current and previous reconstruction frames. However, the main challenge is the lack of labeled data. Recent works such as \cite{DBLP:conf/nips/CaronMMGBJ20, DBLP:conf/icml/ChenK0H20, DBLP:journals/corr/abs-2003-04297} propose contrastive learning mechanisms, which have proven to be powerful tools for learning representations without labeled data. Therefore, we leverage contrastive learning in this paper to address this challenge.

SimCLR~\cite{DBLP:conf/icml/ChenK0H20} introduces a contrastive loss called NT-Xent, which maximizes agreement between augmented views of the same instance. In typical instance discrimination, all instances other than the positive reference are considered negatives. However, in the video task, neighboring frames around the current frame are highly correlated, and regarding them as negatives directly may hinder learning. To address this issue, we propose a novel TCL that minimizes the KL-divergence between the embedding similarity of $\{ \mathbf{\hat{v}}_{t}, \mathbf{\hat{V}}^{p}_{t}\}$ and a prior Gaussian distribution, as shown in Fig. \ref{Contrastive}. Here, $\mathbf{\hat{V}}^{p}_t=\{ \mathbf{\hat{v}}_{t-1},\mathbf{\hat{v}}_{t-2},...\mathbf{\hat{v}}_{t-L}\}$ is the set of previous reconstruction frames from timestamp $t-L$ to $t-1$.

Concretely, we use a pre-trained feature extraction network ResNet~\cite{DBLP:conf/cvpr/HeZRS16} and the global pooling operation to project reconstruction frames $\{ \mathbf{\hat{v}}_{t}, \mathbf{\hat{V}}^{p}_{t}\}$ to latent embeddings $\{ \mathbf{h}_{t}, \mathbf{H}^{p}_t\}$. Due to the fact that temporally adjacent frames are more highly correlated than those faraway ones, we assume the embedding similarity between $\mathbf{h}_{t}$ and $\mathbf{H}^{p}_t$ should follow a prior
Gaussian distribution of timestamp distance between $\mathbf{\hat{v}}_{t}$ and $\mathbf{\hat{V}}^{p}_{t}$ . This assumption motivates us to use KL-divergence to optimize the embedding space. Specifically, let $\operatorname{sim}(\boldsymbol{u}, \boldsymbol{v})=\boldsymbol{u}^{\top} \boldsymbol{v} /\|\boldsymbol{u}\|\|\boldsymbol{v}\|$ denote cosine similarity, and $Gau(x)=\frac{1}{\sigma \sqrt{2 \pi}} \exp \left(-\frac{x^2}{2 \sigma^2}\right)$ denote the Gaussian function, where $\sigma^2$ is the variance. We formulate the loss of the $t$-th frame as:
\begin{equation}
\begin{aligned}
\mathcal{L}_{cont}&=-\sum_{j=t-L}^{t-1} w_{tj} \log \frac{\exp \left(\operatorname{sim}\left(\mathbf{h}_{t}, \mathbf{h}_{j}\right) / \tau\right)}{\sum_{k=t-L}^{t-1} \exp \left(\operatorname{sim}\left(\mathbf{h}_{t}, \mathbf{h}_{k}\right) / \tau\right)}, \\
w_{tj}&=\frac{Gau\left(t-j\right)}{\sum_{k=t-L}^{t-1} Gau\left(t-k\right)},
\end{aligned}
\end{equation}
where $w_{tj}$ is the normalized Gaussian weight between timestamps $t$ and $j$, and $\sigma^2=10$ is as default. $\tau=0.1$ is the temperature parameter. We set $L=80$ in this paper. 

\noindent\textbf{Total Loss.} The total loss $\mathcal{L}_{total}$ in this work is written as:
\begin{equation}
\begin{aligned}
\mathcal{L}_{total} = \mathcal{L}_{spa} + \mathcal{L}_{freq} + \mathcal{L}_{flow} + \mathcal{L}_{ent} +  \mathcal{L}_{cont}.
\end{aligned}
\end{equation}
 
\begin{figure}[!t]
\centering
\includegraphics[width=\linewidth]{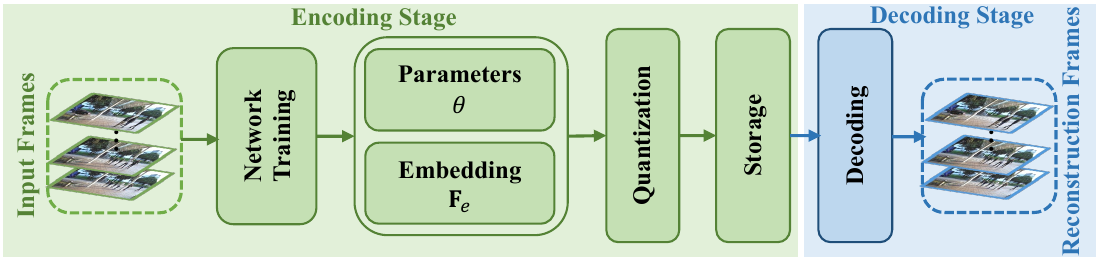}
\caption{Compression pipeline of our proposed MVC method.}
\vspace{-0.5cm}
\label{compression}
\end{figure}

\subsection{Compression Pipeline for Proposed Method}
The compression pipeline of this paper is presented in Fig. \ref{compression}. To compress an input video sequence $\textbf{V}$, at the training (encoding) stage, we first train the proposed network on $\textbf{V}$. The network parameters $\theta$ and the embedding $\mathbf{F}_{e}$ are then quantized into a bitstream for storage and transmission. The network's parameters are single precision floating point numbers that require 32 bits per weight, but to reduce the memory requirement, we use the AI Model Efficiency Toolkit (AIMET)\footnote{\url{https://quic.github.io/}} for quantization. We apply quantization specific to each weight tensor such that the uniformly-spaced quantization grid is adjusted to the value range of the tensor. The bitwidth determines the number of discrete levels, i.e., quantization bins, and we find empirically that bitwidths in the range of 7-8 lead to optimal rate-distortion performance for our models as shown in the supplement material. Apart from AIMET, we can also use the quantization mechanism proposed in COOL-CHIC~\cite{ladune2022cool} to quantize network parameters. Finally, the quantized network parameters and the embedding are used for decoding. Although more advanced model compression techniques, such as model pruning, can further improve performance, the discussion for model compression is beyond the scope of this paper and will be considered in future work.

\subsection{Discussion of Our Method} \label{Discussion}
Our proposed MVC network encodes an entire video sequence simultaneously into network parameters. The bits per pixel (Bpp) for the video sequence are calculated using the formula: $(NP + FE) / (FN \times W \times H)$. Here, \textit{NP} denotes the bits required for network parameters, \textit{FE} denotes the bits required for feature embeddings, and \textit{FN} is the total number of frames, while $W$ and $H$ are the width and height of a frame, respectively.

The current version of the MVC network may only be suitable for the non-live (or non-delay-constrained) scene. In this scene, the video can be encoded and stored first and decoded later for analysis or video on demand (VoD), when required. Unlike traditional video coding standards and DLVC methods that decode video frames sequentially, our MVC network allows for random access to video frames at any time during decoding. This feature enables researchers to easily analyze or request saved videos.

\section{Experiments}

\subsection{Implementation Details}
\noindent\textbf{Experiment Settings.}
We use the PyTorch framework to implement our method, accelerated by the NVIDIA RTX 3090. Following the works~\cite{DBLP:journals/corr/abs-2207-08132,chen2021nerv}, we train the model using Adam optimizer~\cite{DBLP:journals/corr/KingmaB14}. The initialization learning rate is $5e^{-4}$, and decreases 10\% every 10 epochs. Each model is trained with the batchsize of 1. 

\noindent\textbf{Evaluation Metrics.}
In this paper, we use PSNR to measure the quality of the reconstructed frames, which is the commonly used quality metric in video compression. The compression rate is measured by the Bpp. Additionally, we evaluate various video compression methods using the BDBR~\cite{bjontegaard2001calculation}. The BDBR is a measure of how much bit rate is saved when compared to the baseline algorithm at the same quality, measured by PSNR. 

\begin{figure*}[!hbpt]
\centering
\includegraphics[scale=0.78]{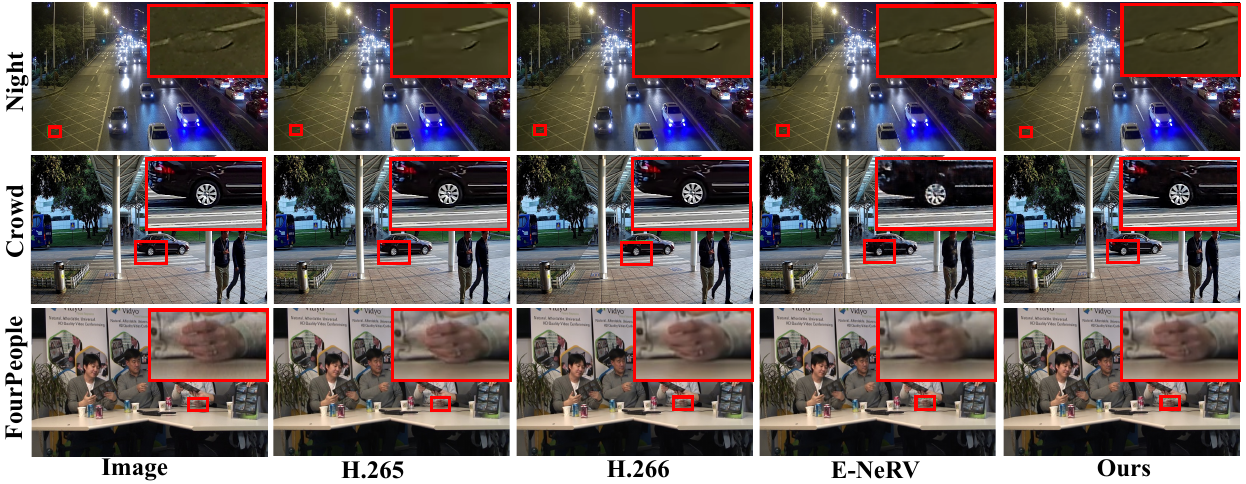}
\caption{The qualitative results of our and other SOTA methods.}
\label{VisResults}
\end{figure*}

\begin{table*}[]
\centering
\caption{BDBR(\%) performances of different methods when compared with H.266 on the 9 different video sequences in terms of PSNR.}
\scalebox{0.72}{
\begin{tabular}{@{}c|cccc|cccc|cccc@{}}
\toprule
                                  & \multicolumn{4}{c|}{\textbf{Surveillance-1}~\cite{ieee1857}}                                                                                                                      & \multicolumn{4}{c|}{\textbf{Surveillance-2}~\cite{oh2011large}}                                                                                                                      & \multicolumn{4}{c}{\textbf{Conference}~\cite{DBLP:journals/tcsv/SullivanOHW12}}                                                                                                                       \\ \cmidrule(l){2-13} 
\multirow{-2}{*}{\textbf{Models}} & \textbf{Crowd}                         & \textbf{Bulong}                        & \textbf{Night}                         & \textit{\textbf{Average}}              & \textbf{Seq001}                        & \textbf{Seq002}                        & \textbf{Seq003}                        & \textit{\textbf{Average}}              & \textbf{FourPeople}                   & \textbf{KristenAndSara}               & \textbf{Johnny}                       & \textit{\textbf{Average}}             \\ \midrule
\textbf{HSTE(MM2022)}             & \textbf{7.12}                          & \textbf{13.43}                         & \textbf{12.25}                         & \textbf{10.93}                         & \textbf{7.65}                          & \textbf{5.43}                          & \textbf{8.87}                          & \textbf{7.32}                          & \textbf{16.75}                        & \textbf{15.83}                        & \textbf{5.46}                         & \textbf{12.68}                        \\
\textbf{CANFVC(ECCV2022)}         & \textbf{8.79}                          & \textbf{12.76}                         & \textbf{14.36}                         & \textbf{11.97}                         & \textbf{8.79}                          & \textbf{9.32}                          & \textbf{7.93}                          & \textbf{8.68}                          & \textbf{15.58}                        & \textbf{16.95}                        & \textbf{7.34}                         & \textbf{13.29}                        \\
\textbf{E-NeRV(ECCV2022)}         & \textbf{10.01}                         & \textbf{17.93}                         & \textbf{13.75}                         & \textbf{13.90}                         & \textbf{12.21}                         & \textbf{11.43}                         & \textbf{8.73}                          & \textbf{10.79}                         & \textbf{35.38}                        & \textbf{18.87}                        & \textbf{28.75}                        & \textbf{27.67}                        \\
\textbf{Ours}                     & {\color[HTML]{FE0000} \textbf{-31.13}} & {\color[HTML]{FE0000} \textbf{-14.94}} & {\color[HTML]{FE0000} \textbf{-22.84}} & {\color[HTML]{FE0000} \textbf{-22.97}} & {\color[HTML]{FE0000} \textbf{-20.97}} & {\color[HTML]{FE0000} \textbf{-16.02}} & {\color[HTML]{FE0000} \textbf{-22.07}} & {\color[HTML]{FE0000} \textbf{-19.68}} & {\color[HTML]{FE0000} \textbf{-8.45}} & {\color[HTML]{FE0000} \textbf{-4.64}} & {\color[HTML]{FE0000} \textbf{-8.83}} & {\color[HTML]{FE0000} \textbf{-7.30}} \\ \bottomrule
\end{tabular}}
\label{BDBR}
\end{table*}

\noindent\textbf{Evaluation Databases.} 
As discussed in Section.\ref{Discussion}, the current version of our proposed MVC network may only be applicable to certain non-live scenes. Therefore, we first evaluate the performance of our proposed method on video sequences that can be first encoded and stored, and then used for analysis or VoD when required, such as conference and surveillance videos. For conference videos, we choose three typical 720p resolution video sequences from HEVC ClassE~\cite{DBLP:journals/tcsv/SullivanOHW12}. For surveillance videos, we choose three 1080p resolution video sequences from IEEE1857~\cite{ieee1857} and three 1080p resolution video sequences from the work~\cite{oh2011large}.

\subsection{Comparison Methods}
We compare our method against traditional codecs, INR-based methods and DLVC approaches. Traditional video compression codecs contain H.265~\cite{DBLP:journals/tcsv/SullivanOHW12} (HM16.21) and H.266~\cite{bross2021overview} (VTM12.0), where H.266 is still a video codec with the best performance for most cases. The INR-based method is E-NeRV~\cite{DBLP:journals/corr/abs-2207-08132} (ECCV2022), which is also similar to our proposed method using implicit neural representation technique. DLVC approachs are HSTE~\cite{DBLP:conf/mm/Li0022} (MM2022) and CANFVC~\cite{ho2022canf} (ECCV2022). HSTE claims that its performance has already surpassed H.266 and it is the industry leader in deep video compression.

\subsection{Quantitative and Qualitative Evaluation}
The quantitative results are presented in Table.~\ref{BDBR}, where we report the BDBR performance. It can be observed that our proposed MVC method consistently outperforms the traditional video compression codec H.266 in all video sequences by a significant margin. Specifically, in the surveillance videos, our method achieves an approximately 1dB PSNR improvement at a similar Bpp compared to H.266, which represents a remarkable performance boost in the video compression task. In Fig.~\ref{VisResults}, we observe that although the SOTA INR-based method E-NeRV is capable of reconstructing the video scene to some extent, its performance is inferior to that of H.266. For instance, in the ``Crowd" sequence, detailed information about the wheel is missing. This illustrates that the existing SOTA INR method fails to capture the spatial context of the current video sequences accurately when given limited network parameters. A straightforward solution for INR networks is to increase the network parameters to enhance the representation capability. However, for the video compression task, we need to focus more on rate-distortion performance. Therefore, we propose CRSPE to enable the INR network to learn from context-related spatial positional embeddings and introduce FDS to capture high-frequency details.

\begin{figure}[!t]
\centering
\includegraphics[width=\linewidth]{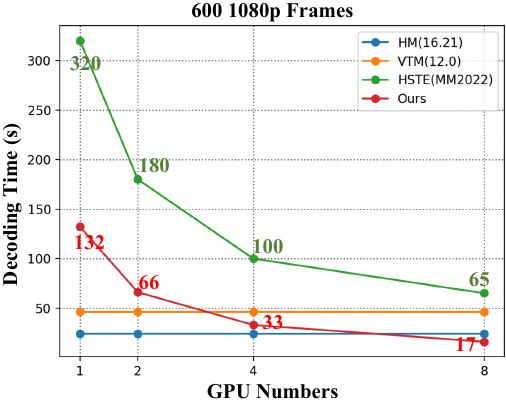}
\caption{Decoding time (s) of our proposed method and other traditional video compression codecs.}
\label{decodingtime}
\vspace{-0.5cm}
\end{figure}

\begin{table*}[!t]
\centering
\caption{The BDBR(\%) performances of different settings when compared with H.266. ``w/o" means ``without" operation.}
\scalebox{0.75}{
\begin{tabular}{@{}c|ccccc|cc|cccc@{}}
\toprule
{\color[HTML]{000000} }                                                                                   & \multicolumn{5}{c|}{{\color[HTML]{000000} \textbf{Architecture}}}                                                                                                                                                                                                                                                                                                       & \multicolumn{2}{c|}{{\color[HTML]{000000} \textbf{SFCM}}}                                                                                                                                        & \multicolumn{4}{c}{{\color[HTML]{000000} \textbf{TCL}}}                                                                                                                                                                            \\ \cmidrule(l){2-12} 
\multirow{-2}{*}{{\color[HTML]{000000} \textbf{\begin{tabular}[c]{@{}c@{}}Video \\ Groups\end{tabular}}}} & {\color[HTML]{000000} \textbf{Baseline}} & {\color[HTML]{000000} \textbf{+CRSPE}} & {\color[HTML]{000000} \textbf{\begin{tabular}[c]{@{}c@{}}+CRSPE\\ +FDS\end{tabular}}} & {\color[HTML]{000000} \textbf{\begin{tabular}[c]{@{}c@{}}+CRSPE\\ +FDS+SFCM\end{tabular}}} & {\color[HTML]{000000} \textbf{\begin{tabular}[c]{@{}c@{}}+CRSPE\\ +FDS+SFCM+TCL\end{tabular}}} & {\color[HTML]{000000} \textbf{\begin{tabular}[c]{@{}c@{}}w/o\\ Regularization\end{tabular}}} & {\color[HTML]{000000} \textbf{\begin{tabular}[c]{@{}c@{}}w/o\\ $\mathcal{L}_{ent}$\end{tabular}}} & \multicolumn{1}{c|}{{\color[HTML]{000000} \textbf{\begin{tabular}[c]{@{}c@{}}w/o\\ PGD\end{tabular}}}} & {\color[HTML]{000000} \textbf{$L=40$}} & {\color[HTML]{000000} \textbf{$L=80$}} & {\color[HTML]{000000} \textbf{$L=120$}} \\ \midrule
{\color[HTML]{000000} \textbf{Surveillance-1}}                                                            & {\color[HTML]{000000} \textbf{13.82}}    & {\color[HTML]{000000} \textbf{3.15}}   & {\color[HTML]{000000} \textbf{-3.53}}                                                 & {\color[HTML]{000000} \textbf{-18.65}}                                                     & {\color[HTML]{000000} \textbf{-22.97}}                                                         & {\color[HTML]{000000} \textbf{-3.65}}                                                        & {\color[HTML]{000000} \textbf{-16.69}}                                                            & \multicolumn{1}{c|}{{\color[HTML]{000000} \textbf{-18.12}}}                                            & {\color[HTML]{000000} \textbf{-19.85}} & {\color[HTML]{000000} \textbf{-22.97}} & {\color[HTML]{000000} \textbf{-21.97}}  \\ \midrule
{\color[HTML]{000000} \textbf{Surveillance-2}}                                                            & {\color[HTML]{000000} \textbf{10.74}}    & {\color[HTML]{000000} \textbf{3.34}}   & {\color[HTML]{000000} \textbf{-3.61}}                                                 & {\color[HTML]{000000} \textbf{-15.67}}                                                     & {\color[HTML]{000000} \textbf{-19.68}}                                                         & {\color[HTML]{000000} \textbf{-3.68}}                                                        & {\color[HTML]{000000} \textbf{-13.47}}                                                            & \multicolumn{1}{c|}{{\color[HTML]{000000} \textbf{-15.52}}}                                            & {\color[HTML]{000000} \textbf{-17.03}} & {\color[HTML]{000000} \textbf{-19.68}} & {\color[HTML]{000000} \textbf{-18.86}}  \\ \midrule
{\color[HTML]{000000} \textbf{Conference}}                                                                & {\color[HTML]{000000} \textbf{27.59}}    & {\color[HTML]{000000} \textbf{14.24}}  & {\color[HTML]{000000} \textbf{3.21}}                                                  & {\color[HTML]{000000} \textbf{-4.21}}                                                      & {\color[HTML]{000000} \textbf{-7.30}}                                                          & {\color[HTML]{000000} \textbf{4.33}}                                                         & {\color[HTML]{000000} \textbf{-3.31}}                                                             & \multicolumn{1}{c|} {\color[HTML]{000000} \textbf{-3.97}}                                                                  & {\color[HTML]{000000} \textbf{-5.04}}  & {\color[HTML]{000000} \textbf{-7.30}}  & {\color[HTML]{000000} \textbf{-6.15}}   \\ \bottomrule
\end{tabular}}
\label{ablation}
\end{table*}

Another limitation of the existing SOTA INR-based methods is that they struggle to fully capture the temporal correlation of video sequences. For example, when reconstructing regions with complex motions, such as finger regions in the ``FourPeople" sequence, there are noticeable blurs. To address this issue, we have designed the SFCM and TCL modules, which are specifically aimed at making INR networks more effective for complex scenes. As shown in Fig. \ref{VisResults} and Table. \ref{BDBR}, our proposed MVC method is better able to handle conference scenes with complex motions compared to E-NeRV. Note that, to align the bitrate points of our method and HSTE and CANFVC, we retrain these two methods using the source code provided by the authors.

\subsection{Decoding Time Comparison}
We analyze our method's decoding time. The platform is Intel(R) Xeon(R) Gold 5218R CPU and NVIDIA RTX3090 GPU. The results are shown in Fig. \ref{decodingtime}. When decoding 600 1080p video frames, HM16.21 takes approximately 24 seconds, and VTM12.0 takes about 47 seconds. With one GPU, our proposed method takes around 132 seconds, while HSTE takes about 320 seconds to decode the same number of frames. However, as the number of GPUs increases, our proposed method becomes faster and finally achieves real-time decoding at 35 frames per second (FPS). Our proposed method utilizes a network that directly models the entire 600 frames, making it parallel-friendly and capable of utilizing all GPUs to their full potential. In contrast, HSTE decodes frames using a sequential process, making it parallel-unfriendly and unable to fully utilize all GPUs. Note that, our network allows for random access to video frames at any time during decoding. This feature enables researchers to easily analyze or request saved videos.

\subsection{Ablation Studies}
\textbf{About the whole architecture.} In this paper, we mainly propose four modules to improve the performance, including CRSPE, FDS, SFCM and TCL. We progressively integrate these modules into the \textit{\textbf{Baseline}} network, and the performance change is shown in Table. \ref{ablation} (\textit{\textbf{Architecture}}). The architecture of \textit{\textbf{Baseline}} is similar to the gray regions in Fig. \ref{framework} (Without ``\textit{C}" operation). It can be seen that all these four modules can consistently improve the network performance. The series of experiments validate the effectiveness of our proposed modules in this paper.

\textbf{About SFCM.} In the proposed SFCM, we design two important mechanisms, including flow regularization and $\mathcal{L}_{ent}$. Their ablation studies are shown in Table. \ref{ablation} (\textit{\textbf{SFCM}}). Firstly, without flow regularization operation \textit{\textbf{(SFCM(w/o Regularization))}}, the performance of our proposed network has no obvious improvement, compared to the (\textit{\textbf{+CRSPE+FDS}}). Regretfully, we even find that there is a bit of performance loss in the Conference group. The main reason is that without a flow regularization mechanism, the supervised labels would contain noise. The noise information would disrupt the whole training process. Moreover, we further design the $\mathcal{L}_{ent}$ to constrain the regularization map $\mathbf{W}$ (Eqn. \ref{ent}). Without $\mathcal{L}_{ent}$, the $\mathbf{W}$ can be viewed as the one-channel attention map, and the final reconstruction results of the whole MVC network would indirectly supervise the $\mathbf{W}$. Hence, there exist the risks that the $\mathbf{W}$ cannot be well learned, leading to network performance (\textit{\textbf{SFCM(w/o $\mathcal{L}_{ent}$)}}) drop to some extent, compared to the (\textit{\textbf{+CRSPE+FDS+SFCM}}). Adding $\mathcal{L}_{ent}$, it directly supervises the regularization map $\mathbf{W}$ and makes the map tend to be binary, which lets the network be converged better.

\textbf{About TCL.} In our proposed TCL, we design the prior
Gaussian distribution (PGD) to ensure the TCL can correctly select positive samples. As shown in Table. \ref{ablation} (\textit{\textbf{TCL}}), without PGD (\textit{\textbf{w/o PGD}}), the proposed contrastive loss has no contribution to long-term temporal correlation modeling. Moreover, we also find that the length variable $L$ of selected previous frames would affect the performance of TCL. Compared with larger values $L$=80 and 120, the smaller value would slightly hurt the performance, but it still can significantly improve the performance compared to the model without using TCL (\textit{\textbf{+CRSPE+FDS+SFCM}}). In this paper, we set $L=80$, since we find that a longer $L$ cannot further improve the performance.

\section{Conclusion}
In this paper, we propose a novel MVC framework for the video compression task. We leverage the INR network as our backbone network, and discuss the limitations of existing INR networks when they are applied to the video coding task. To address these limitations, we propose context-related spatial positional embedding and frequency domain supervision to enhance the spatial context ability of existing INR networks. Moreover, we design the scene flow constrain mechanism and temporal contrastive loss to improve the temporal modeling ability. In experiments, our proposed MVC method consistently outperforms H.266 for all the test video sequences, which may inspire researchers to explore the video compression task in a new light.

{\small
\bibliographystyle{ieee_fullname}
\bibliography{egbib}
}

\end{document}